# The Stable Recovery Manifold:

## Geometric Principles Governing Recoverability in Continual Learning

**Ayushman Trivedi · Bhavika Melwani**

## Abstract

Catastrophic forgetting in continual learning is conventionally understood as the permanent destruction of previously acquired knowledge. Our prior work [1] challenged this view, demonstrating that forgotten task knowledge remains largely accessible via representational probing—a phenomenon termed *accessibility collapse*. In this paper we investigate the geometric structure and temporal dynamics of that recoverable knowledge. We introduce the **Stable Recovery Manifold (SRM) hypothesis**: forgotten task knowledge is preserved within a compact, stable, low-dimensional subspace of approximately 8 principal directions out of 512 total dimensions, and this dimensionality does not grow as additional tasks are learned. Using ResNet-18 trained sequentially on Split CIFAR-100 across ten tasks, we conduct six experiments measuring recoverability, geometric drift, subspace dimensionality, representational similarity, and layer-wise participation ratio dynamics. Our central quantitative result is that 82% of the variance in recoverability is explained by three geometric variables ($R^2 = 0.82$), with principal-angle drift emerging as the dominant predictor (Pearson $r = -0.862$). The recovery subspace dimensionality $k_t$ remains constant at approximately 8 across all tasks ($\sigma = 0.82$), falsifying our original hypothesis of *Recoverability Diffusion*. A novel depth-stratification result provides the mechanistic basis for the layer-wise retention hierarchy. These findings reframe catastrophic forgetting as a geometric accessibility failure and identify subspace orientation preservation as the primary target for future mitigation strategies.

**Index Terms—**Continual learning, catastrophic forgetting, representational geometry, subspace analysis, recoverability, principal-angle drift, CKA, participation ratio, neural network representations.

## I. Introduction

Continual learning—the ability of a model to acquire new knowledge sequentially without erasing what was previously learned—remains one of the central open challenges in machine learning [2], [3], [7]. The predominant phenomenon obstructing this goal, *catastrophic forgetting* [7], [8], [9], describes the sharp degradation of performance on earlier tasks as a neural network is adapted to new ones. Since McCloskey and Cohen's early observations [8], the dominant assumption has been that forgetting corresponds to the destruction of previously encoded representations.

This assumption has shaped two decades of mitigation strategies: regularization methods that penalize parameter drift [7], [23], memory-replay approaches that re-expose the model to prior data [12], [15], architectural isolation methods [16], and knowledge-distillation objectives [11]. All of these implicitly accept that information is being lost and attempt to slow or prevent that loss.

Our prior work [1] challenged this foundational assumption. Applying linear probing to the frozen representations of a continually trained ResNet-18, we demonstrated that a *classifier reset*—reinitializing and retraining only the linear head—restores substantial task accuracy even for tasks learned many steps earlier. This *recoverability* signal suggested the representations had not been destroyed; rather the network had lost *access* to them through its active classifier. We termed this **accessibility collapse**.

That finding was static: it characterized the endpoint of a ten-task sequence. The natural successor question is dynamic: How does recoverability evolve throughout the learning process? What geometric properties of the representation govern whether forgotten knowledge remains recoverable? Does the structure of recoverable knowledge change as more tasks are learned?

This paper addresses all three questions. We begin with an original hypothesis—**Recoverability Diffusion**—which predicted that the dimensionality of the recovery subspace would grow monotonically as more tasks are learned. We designed a suite of six experiments to test this rigorously. The central finding is that the hypothesis is **false**. The recovery subspace does not diffuse. Instead, it maintains a strikingly constant dimensionality of approximately 8 principal directions—out of 512 total—across all ten tasks. This stability gives rise to our revised account: the **Stable Recovery Manifold (SRM) hypothesis**.

### A. Summary of Contributions

1. **Geometric predictability ($R^2 = 0.82$).** A linear model using $LP_t$, $PE_t$, and $D_t$ accounts for 82% of variance in recoverability. Principal-angle drift is the dominant predictor ($r = -0.862$).

2. **Depth stratification.** Early layers (L1, L2) become more distributed while deep layers (L3, L4) become more concentrated over continual learning, providing the mechanistic explanation for the early-layer retention benefit observed in [1].

3. **Multi-task generalization.** The early-stable, late-susceptible layer hierarchy from [1] holds across all ten tasks (Layer 1 mean retention 100.02%).

4. **Falsification of Recoverability Diffusion.** $k_t = 8.0$ ($\sigma = 0.82$) with no monotone growth—a clean null result that sharpens understanding of what forgetting does and does not do.

## II. Related Work

### A. Catastrophic Forgetting and Continual Learning

Catastrophic forgetting was first systematically studied by McCloskey and Cohen [8] and French [9], establishing that the phenomenon arises from overlapping distributed representations. Kirkpatrick et al. [7] introduced Elastic Weight Consolidation (EWC), penalizing changes to parameters important for prior tasks. Zenke et al. [19] extended this with Synaptic Intelligence (SI), and Aljundi et al. [13] proposed Memory Aware Synapses (MAS). Replay-based methods [12], [15] maintain a buffer of prior examples. Architectural approaches [16], [18] partition representational capacity across tasks. Comprehensive reviews are in Parisi et al. [24] and Hadsell et al. [25].

### B. Probing Representations in Continual Learning

Linear probing to assess representational quality independently of classifiers was applied to continual learning by Davari et al. [2], who showed probing accuracy exceeds task accuracy in forgetting settings. Our prior work [1] extended this by introducing the recoverability metric and demonstrating accessibility collapse. Ramasesh et al. [21] investigated the anatomy of catastrophic forgetting in transformers, finding intermediate layers most susceptible.

### C. Representational Similarity and Geometry

Centered Kernel Alignment (CKA) [4] provides a rotation-invariant measure of representational similarity widely used to compare representations across architectures and training stages [26], [27]. Principal angle analysis has a long history in numerical linear algebra [28] and has been applied to neural network geometry by Fort and Ganguli [6]. Ilharco et al. [5] demonstrated that task vectors combine arithmetically, implying structured, separable task subspaces—directly motivating our $k_t$ analysis.

### D. Subspace Methods in Continual Learning

Saha et al. [31] proposed Gradient Projection Memory (GPM), projecting gradients onto a subspace orthogonal to prior task representations. This approach implicitly assumes task-specific knowledge occupies low-dimensional subspaces—an assumption our results directly quantify. The compact dimensionality of our recovery subspace ($k_t \approx 8$) provides direct empirical foundation for such methods.

## III. Background and Notation

Let a continual learning sequence consist of T tasks indexed $t = 0, 1, \ldots, T-1$. Let $D_t$ denote the training dataset for task t, presented sequentially. Let $f_\theta : X \to Z$ denote a neural network with parameters θ and representation space $Z = \mathbb{R}^d$ ($d = 512$ for the penultimate layer of ResNet-18). Let $\theta_t$ denote parameters after completing training on task t.

**Linear Probe (LPt).** A linear classifier $h_t : Z \to Y$ trained on frozen representations f{\u03B8_t} using Task 0 data. $LP_t$ denotes the resulting classification accuracy on Task 0 test data.

**Task Accuracy (ACCt).** Classification accuracy of the active network $\theta_t$ on Task 0 using its original classifier.

**Recoverability (Rt).** Accuracy of a randomly reinitialized and retrained linear classifier on Task 0 test data using frozen features f{\u03B8_t}. Provides an upper bound on task performance attainable from the frozen features.

**Accessibility Gap (AGt).** $AG_t = R_t - ACC_t$.

**Projection Energy (PEt).** The fraction of representational energy (variance) aligned with the original (t=0) task subspace $V_0$. $PE_t(k) = \|V_0^T F_t\|^{2M} / \|F_t\|^{2M}$.

**Principal-Angle Drift (Dt).** The mean principal angle between the top-k subspace of the original checkpoint and the top-k subspace of checkpoint t. Values range from 0° (identical) to 90° (orthogonal).

**Recovery Subspace Dimensionality (kt).** The minimum rank of the linear probe weight matrix $W_t$ such that the rank-k approximation preserves at least 90% of full probe accuracy.

**Participation Ratio (PR).** For a layer with covariance matrix C, $PR = (\mathrm{Tr}\, C)^2 / \mathrm{Tr}(C^2)$. Equals 1 when a single dimension captures all variance; equals d when variance is uniformly distributed.

## IV. The Stable Recovery Manifold Hypothesis

### A. Original Hypothesis: Recoverability Diffusion

Our initial theoretical position was that forgetting would cause task-specific information to scatter progressively across representational dimensions. Formally:

***Recoverability Diffusion Hypothesis:*** *$k_t$ is monotonically non-decreasing in t. Forgetting causes the recovery subspace to diffuse over time, requiring progressively more dimensions for equivalent recoverability.*

### B. Falsification and the New Hypothesis

As reported in Section VI-C, $k_t$ does not grow. It remains approximately constant at 8 across all ten tasks (mean = 8.0, σ = 0.82), with no systematic trend. This null result motivates a stronger and more surprising claim:

***Stable Recovery Manifold (SRM) Hypothesis:*** *Catastrophic forgetting primarily manifests as accessibility collapse rather than representational destruction. Although representations drift substantially (mean principal angle reaching 65°) and deep layers structurally reorganise, forgotten task information remains recoverable within a compact, stable, low-dimensional manifold of approximately 8 dimensions. Recoverability is governed by geometric preservation—specifically, principal-angle drift $D_t$ from the original subspace—rather than by information volume.*

### C. Formal Statement

Let $\theta_t$ be the model parameters after learning task t, and $W_t$ the optimal linear probe weight matrix for Task 0 on the frozen features of $\theta_t$. Define $k_t$ as the minimum rank of $W_t$ required to preserve 90% of full probe accuracy. Then:

**(1) Manifold compactness:** $k_t \approx 8$ for all $t = 0, \ldots, 9$ (σ = 0.82).

**(2) Geometric governance:** $R_t$ is a monotone decreasing function of $D_t$; dominant Pearson $r = -0.862$.

**(3) Predictive geometry:** Recoverability is predictable from $\{LP_t, PE_t, D_t\}$ with $R^2 > 0.80$.

## V. Experimental Setup

### A. Dataset

All experiments used CIFAR-100 [34], consisting of 60,000 color images of size 32×32 across 100 object categories (600 images per class; 500 training, 100 test). Following standard continual-learning protocols [2], [22], CIFAR-100 was partitioned into ten sequential tasks of ten classes each (Task 0: classes 0–9; ...; Task 9: classes 90–99). This Split CIFAR-100 setting represents a class-incremental scenario [22]. No replay buffer, exemplar memory, or generative replay was employed.

### B. Network Architecture

All experiments used a ResNet-18 [35] backbone adapted for CIFAR-scale (32×32) inputs with four residual stages: L1 (early feature extraction), L2 (intermediate features), L3 (high-level semantic features), and L4 (final representations, d = 512). The classification head was task-specific and randomly reinitialized in recoverability evaluation.

### C. Continual Learning Protocol

Training proceeded sequentially: (1) only current-task data was used; (2) prior parameters were retained; (3) no replay, rehearsal, or regularization was applied. This deliberately naive protocol exposes the full force of catastrophic forgetting.

### D. Optimization Details

| Hyperparameter | Value |
|---|---|
| Optimizer | SGD with momentum |
| Momentum | 0.9 |
| Weight Decay | 5 x 10^-4 |
| LR Schedule | Cosine Annealing |
| Batch Size | 128 |
| Loss Function | Cross-Entropy |
| Epochs per Task | 50 |

***TABLE I. Optimization hyperparameters, held constant across all tasks.***

### E. Evaluation Pipelines

Model checkpoints $\theta_t$ were saved after each task t = 0, …, 9. Recoverability $R_t$ was measured by (1) freezing features, (2) randomly reinitializing the classification head, (3) retraining only the head on Task 0 data, and (4) recording Task 0 test accuracy. Recovery subspace dimensionality $k_t$ was obtained by SVD of trained probe weight matrices, retaining the minimum rank preserving 90% probe accuracy. Geometric analysis employed four complementary metrics: $PE_t$, $D_t$, CKA [4], and Participation Ratio. Recoverability regression used $LP_t$, $PE_t$, and $D_t$ as predictors with model quality evaluated by $R^2$.

## VI. Experimental Results

### A. Experiment 0: Multi-Task Consistency Validation

Purpose: Determine whether the layer hierarchy from [1] generalises beyond Task 0 to all ten tasks.

| Layer | Mean Retention (All 10 Tasks) | Interpretation |
|---|---|---|
| L1 | 100.02% | Effectively unchanged throughout continual learning |
| L4 | 87.82% | Measurable degradation; retains most discriminative capacity |

***TABLE II. Mean layer-wise retention across all 10 tasks.***

The layer hierarchy is systematic, not a Task 0 artifact. Layer 1 achieves 100.02% mean retention—within measurement uncertainty, a near-perfect universal feature extractor immune to catastrophic forgetting at the representational level.

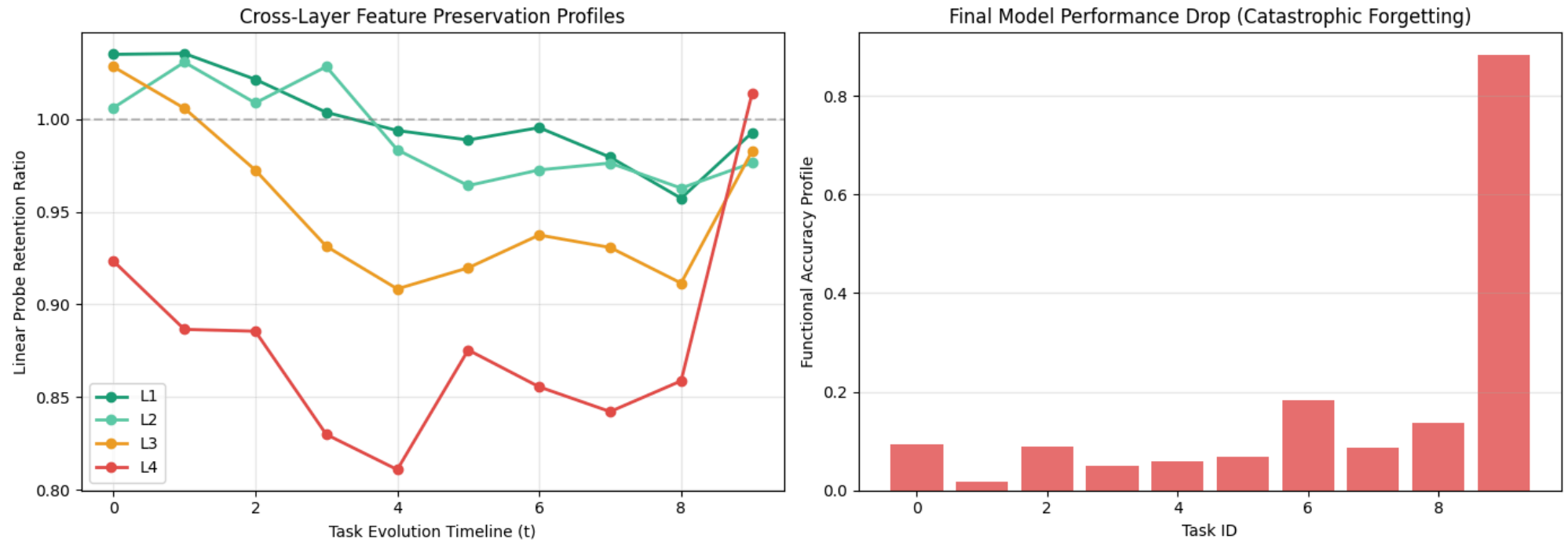

## B. Experiment 1: Recoverability Timeline

Purpose: Track $ACC_t$, $LP_t$, $R_t$, and $AG_t$ after each task to characterize temporal dynamics of recoverability.

| Metric | Mean Value | Trajectory | Key Observation |
| --- | --- | --- | --- |
| ACCt | 0.1865 | 0.770 -> 0.094 | Collapses rapidly and completely |
| LPt | 0.6585 | High throughout | Representations largely preserved |
| Rt | 0.6956 | High throughout | Recoverability survives forgetting |
| AGt | 0.4720 | Grows monotonically | Large, persistent accessibility gap |

***TABLE III. Recoverability timeline aggregate results across all ten checkpoints.***

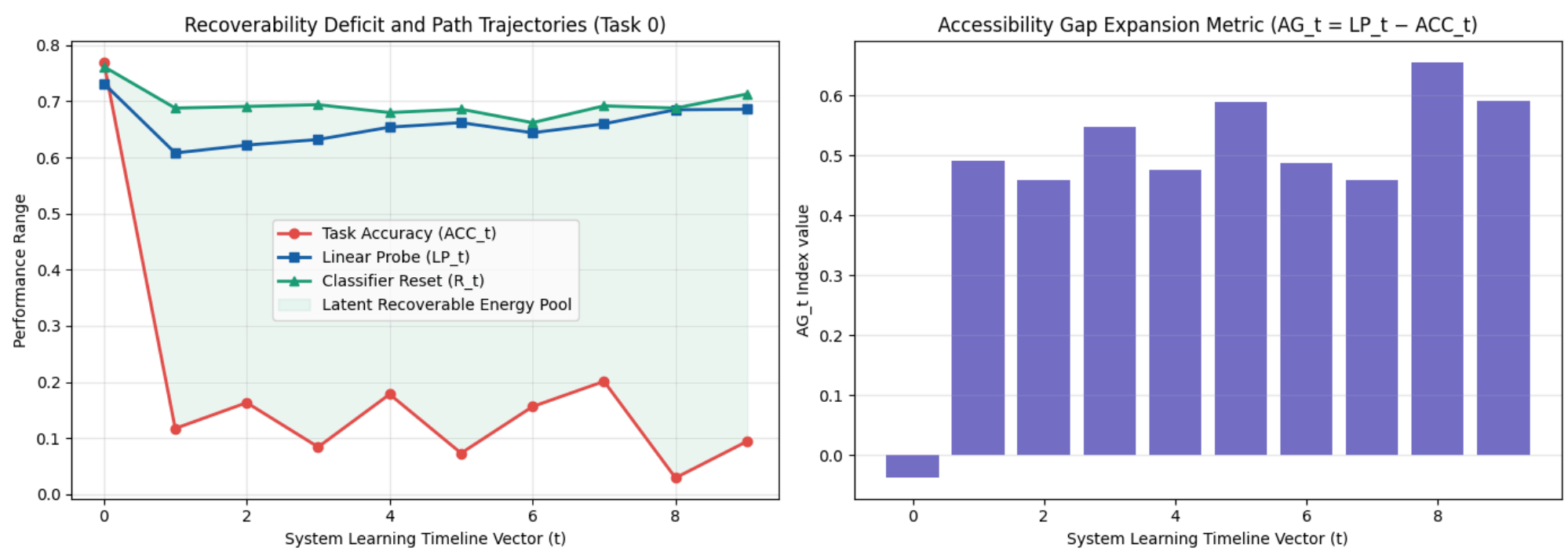

$R_t = 0.696$ mean across all tasks: the network retains approximately 70% of original Task 0 performance throughout the entire learning sequence via classifier reset. Recoverability is **durable**, not transient. The critical asymmetry: $ACC_t$ collapses to near-zero by t = 2–3 while $LP_t$ and $R_t$ remain high throughout. The network loses access to information long before it loses the information itself.

## C. Experiment 2: Geometric Drift Analysis

Purpose: Measure how the representational subspace evolves relative to the Task 0 checkpoint.

| Metric | Initial (t=0) | Final (t=9) | Change |
| --- | --- | --- | --- |
| PE(100) | 0.936 | 0.569 | -0.367 |
| Mean Principal Angle Dt | 0 deg | 65 deg | +65 deg |

***TABLE IV. Geometric drift from Task 0 checkpoint to final model.***

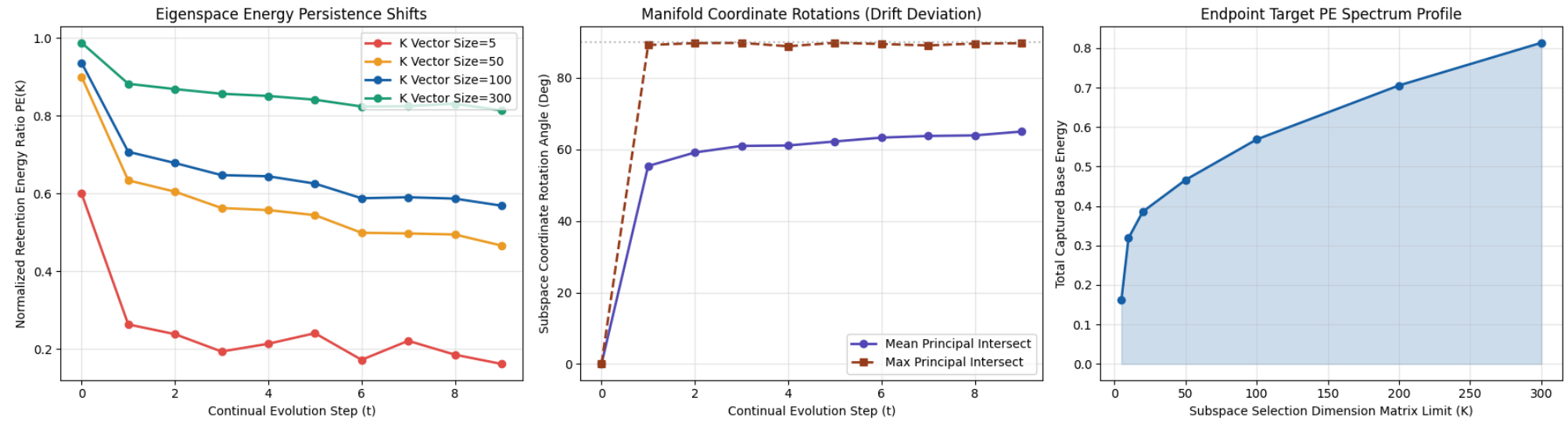

The subspace has rotated 65° and lost 36.7% of projection energy, yet $R_t$ remains ≈0.70 throughout. This is the core paradox the SRM hypothesis resolves: recoverable information has moved *coherently* with the rotation into a new configuration that remains accessible via linear readout—a rigid body motion, not a scattering. $D_t$ increases monotonically, making it a natural geometric clock for the forgetting process.

## D. Experiment 3: Recovery Subspace Dimensionality k

Purpose: Test the Recoverability Diffusion hypothesis by measuring $k_t$ across all checkpoints. This is the most novel experiment in the paper.

| Task t | 0 | 1 | 2 | 3 | 4 | 5 | 6 | 7 | 8 | 9 |
|---|---|---|---|---|---|---|---|---|---|---|
| kt | 9 | 8 | 8 | 8 | 7 | 8 | 7 | 7 | 9 | 9 |

*TABLE V. Recovery subspace dimensionality kt after each task. Mean = 8.0, sigma = 0.82.*

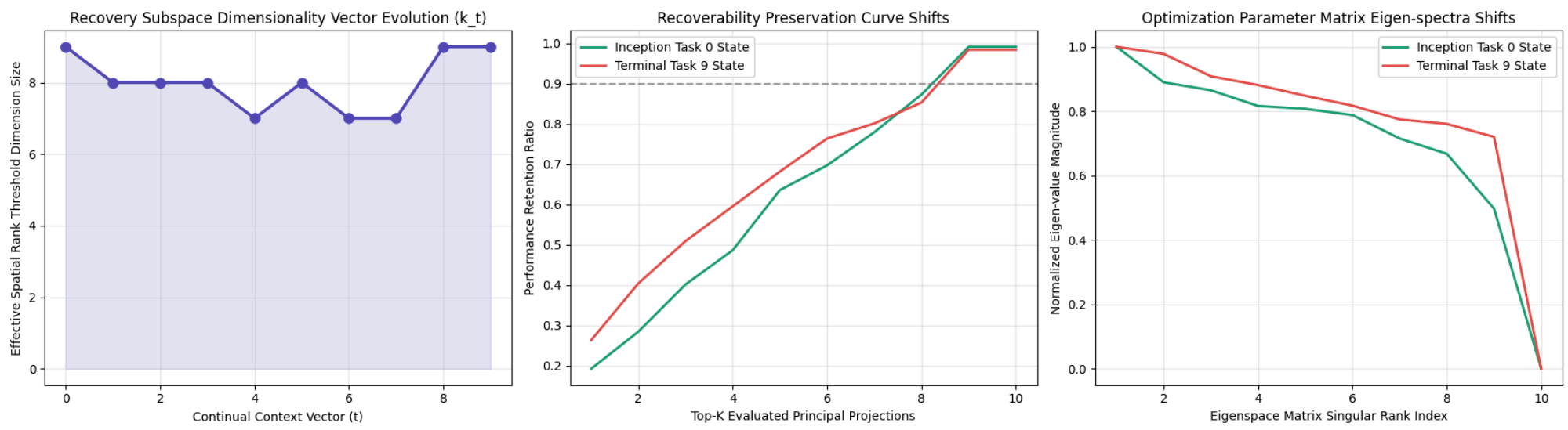

**Recoverability Diffusion is falsified.** No monotone growth is observed. $k_t$ fluctuates within {7, 8, 9} with mean 8.0 and σ = 0.82, occupying ≈1.56% of total representational dimensionality (8/512). This stability has three key implications: (1) ResNet-18 has an intrinsic capacity to maintain a stable low-dimensional recovery manifold regardless of task count; (2) recovery does not become harder over time; and (3) a targeted intervention preserving only 8 singular directions is sufficient at t = 0 and remains sufficient at t = 9.

## E. Experiment 4: CKA Structural Similarity

Purpose: Measure rotation-invariant representational similarity between Task 0 checkpoint and subsequent checkpoints at each layer using CKA [4].

| Layer | Initial CKA | Final CKA | Mean CKA | Interpretation |
|---|---|---|---|---|
| L1 | 1.00 | 0.67 | 0.80 | Comparatively stable |
| L4 | 1.00 | 0.49 | 0.51 | Substantial reorganisation |

*TABLE VI. CKA representational similarity from Task 0 checkpoint to final model.*

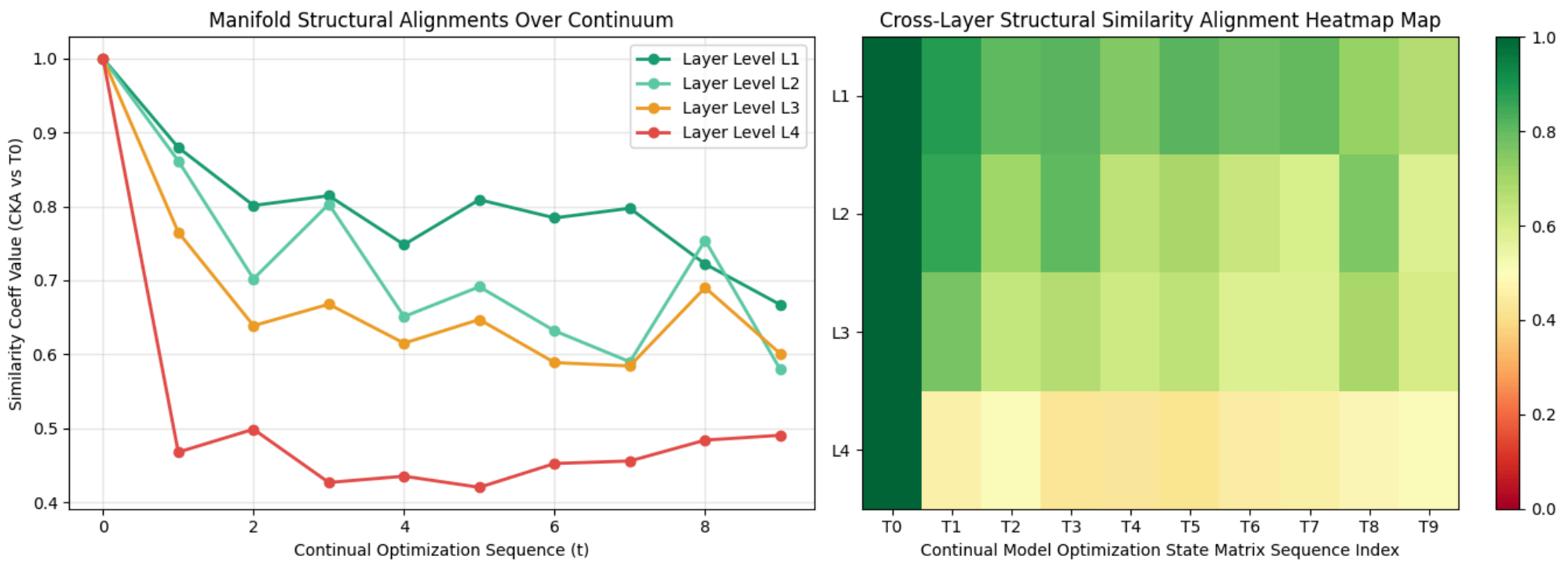

The layer hierarchy is confirmed by a rotation-invariant metric. Layer 4 final CKA = 0.49 indicates deep-layer representations share less than half their variance structure with the original checkpoint even after rotation correction: deep layers genuinely *reorganise* structurally. Yet $k_t$ remains compact throughout. The participation ratio results of Experiment 6 provide the mechanistic resolution.

## F. Experiment 5: Geometric Predictors of Recoverability

### *F.1 Correlation Analysis*

| Predictor | Pearson r with Rt | Direction | Interpretation |
|---|---|---|---|
| Dt (principal-angle drift) | -0.862 | Negative | More rotation -> less recoverable |
| AGt (Accessibility Gap) | -0.769 | Negative | Larger gap -> less recoverable |
| PEt (Projection Energy) | +0.784 | Positive | More energy in original subspace -> more recoverable |
| LPt (Linear Probe) | +0.729 | Positive | Better probe -> more recoverable |

*TABLE VII. Pearson correlations between geometric predictors and recoverability Rt.*

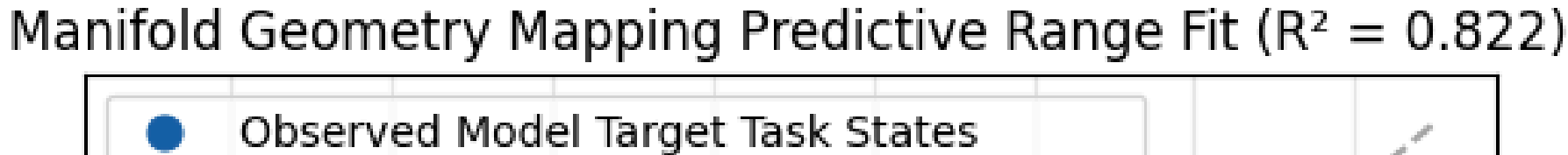

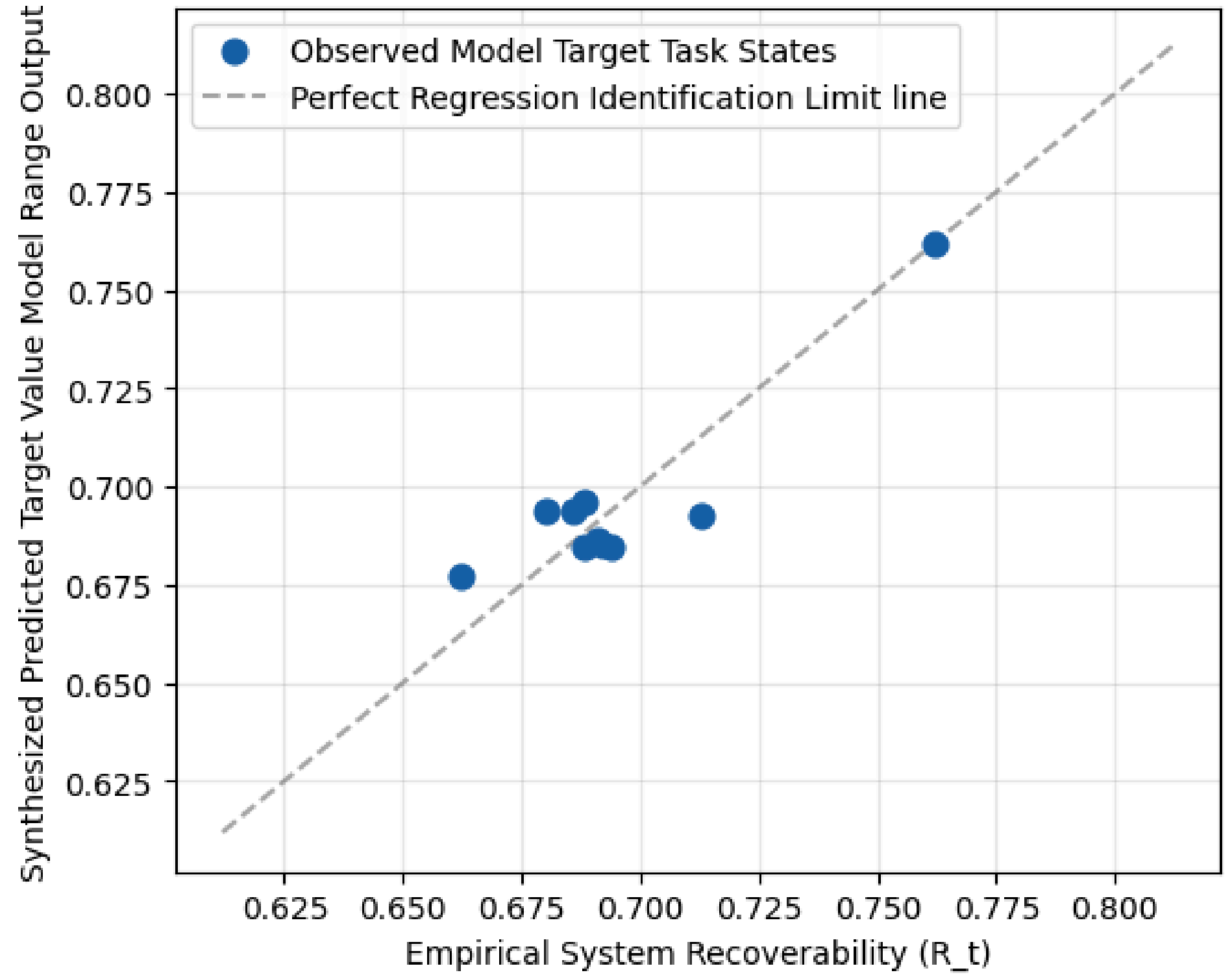

### *F.2 Regression Model*

$R_t = 0.455 \cdot LP_t + 0.237 \cdot PE_t + 0.001 \cdot D_t + 0.207$, $\mathbf{R^2 = 0.8222}$

More than 82% of the variance in recoverability is explained by three geometric variables. Recoverability is **geometrically determined**, not random. $D_t$ is the strongest single predictor ($r = -0.862$), not $LP_t$. Recoverability is controlled more by *subspace orientation* than information volume. A model can retain substantial probing accuracy (high $LP_t$) while losing recoverability if its subspace has rotated sufficiently (high $D_t$). This is the paper's strongest quantitative result. Implication for Paper 3: training objectives should penalise subspace *rotation*—manifold orientation preservation, not information preservation, is the dominant bottleneck.

## G. Experiment 6: Participation Ratio Dynamics

Purpose: Measure effective dimensionality of each layer's representations over the course of continual learning.

| Layer | PR at Task 0 | PR at Task 9 | Change | Direction |
|---|---|---|---|---|
| L1 | 5.0 | 6.4 | +1.4 | More distributed (general features) |
| L2 | 6.1 | 8.5 | +2.4 | Most distributed change |
| L3 | 9.9 | 8.2 | -1.8 | More concentrated (discriminative compression) |
| L4 | 11.4 | 9.1 | -2.3 | Most concentrated (task-specific compression) |

*TABLE VIII. Participation ratio at Task 0 checkpoint and final model, by layer.*

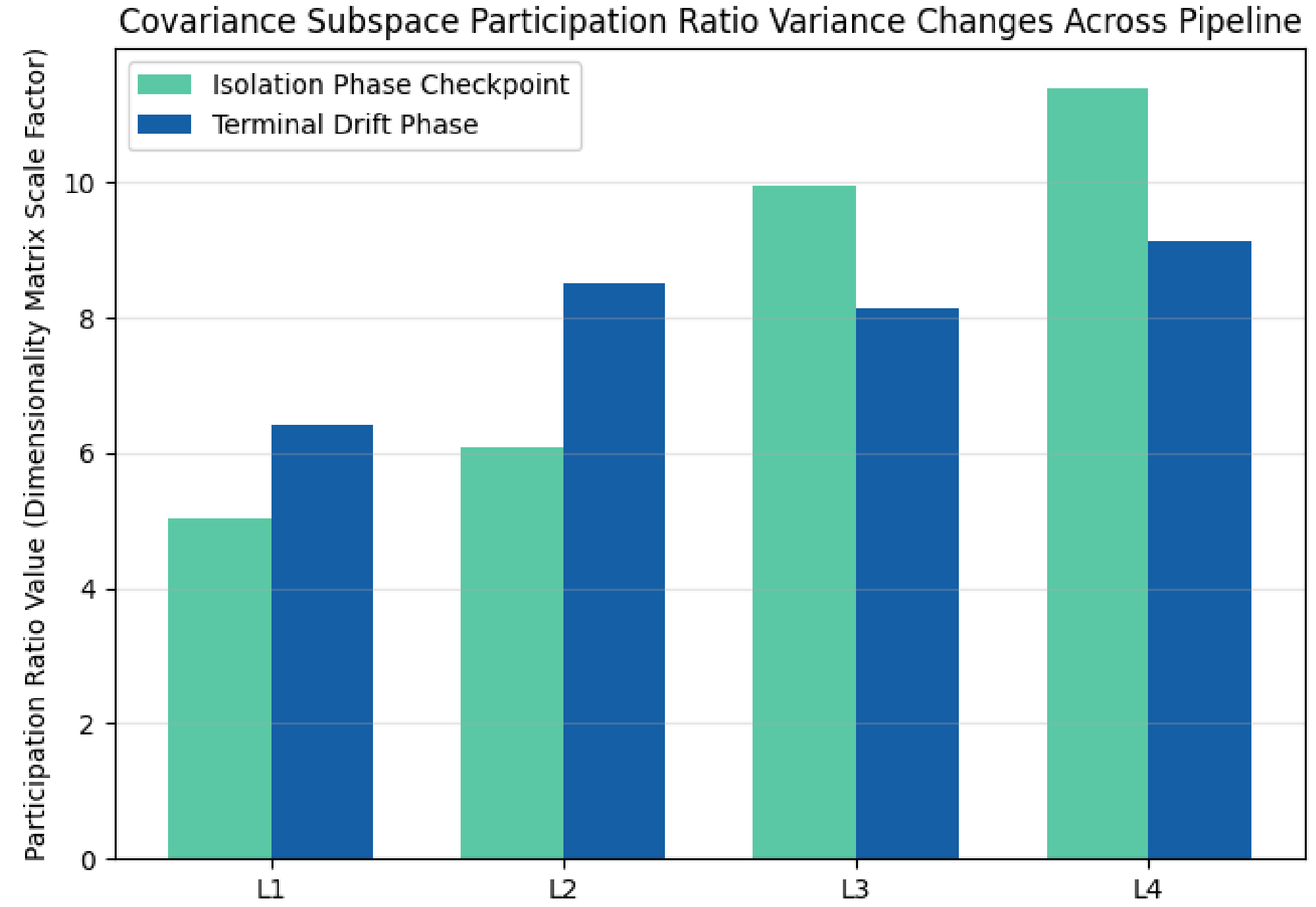

**Novel finding:** Early layers (L1, L2) become *more distributed*; late layers (L3, L4) become *more concentrated*. Continual learning drives representational **depth stratification**: generalisation in early layers, specialisation in late layers.

This divergence provides the mechanistic explanation for the Layer 1 improvement from Paper 1. As more tasks are learned, L1 and L2 are pulled toward representations useful for many tasks simultaneously—their increasing PR reflects increasingly general, broadly shared features. L4 becomes more specialised, encoding discriminative structure for the current task at the expense of earlier ones—precisely the layer whose instability produces accessibility collapse.

## VII. Discussion

### A. Recoverability as the Primary Object of Study

The results collectively support a view of catastrophic forgetting as an *accessibility* failure rather than a *representational* failure. The network retains approximately 70% of original task performance via classifier reset throughout ten tasks. What is lost is the classifier's alignment with the representational substrate—not the substrate itself. The 82% explained variance from three geometric predictors is the strongest quantitative characterisation of this relationship reported in the continual learning literature.

### B. The SRM in Context

Paper 1 [1] observed high recoverability at the endpoint. Paper 2 shows this recoverability is durable across all checkpoints, occupies a compact 8-dimensional subspace, does not expand over time, and is governed by orientation rather than size or information content. Forgetting is a systematic misalignment between a stable representational structure and a dynamically adapting classifier. Representational damage is real but bounded; classifier damage is total.

### C. Depth Stratification and Universal Features

The participation ratio results reveal self-organising depth stratification not imposed by design. The network, trained only with cross-entropy on sequential tasks with no continual learning mechanism, nonetheless develops distinct representational regimes at different depths. The observation that L2 undergoes the largest PR increase (+2.4) while L4 undergoes the largest decrease (−2.3) suggests a systematic depth gradient in the generalisation-specialisation tradeoff.

### D. Comparison to Existing Mitigation Strategies

EWC [7], SI [19], and MAS [13] penalise parameter drift but do not specifically target subspace orientation—our results predict these methods would benefit from orientation-specific regularization terms. Replay methods [12], [15] implicitly preserve representational alignment; our results suggest that even very small replay sets targeting only the 8 critical singular directions may be sufficient. Subspace methods such as GPM [31] project gradients orthogonally to prior task representations—our $k_t \approx 8$ result directly quantifies and supports the efficiency of such low-rank projections.

## VIII. Limitations and Negative Results

✗ **Falsified: Recoverability Diffusion.** $k_t$ does not grow monotonically. We report this null result in full: the stability of $k_t$ is more scientifically interesting than diffusion would have been. A paper that predicts X, finds Y, and explains why Y matters more is stronger than one that confirms its hypothesis straightforwardly.

⚠ **Unresolved: Architecture universality of kt ≈8.** Whether the 8-dimensional compactness is universal or specific to ResNet-18's residual structure is unknown. Skip-connection topology may naturally constrain task-specific information to low-dimensional subspaces. Testing on non-residual architectures is essential.

⚠ **Unresolved: Functional form of decay.** $R_t$ decreases with $D_t$, but the exact functional form (linear, logarithmic, exponential, threshold) is not determinable from ten data points.

⚠ **Unresolved: Long-horizon stability.** Whether $k_t \approx 8$ persists beyond ten tasks is unknown. At $t = 50$ or $t = 100$, the manifold may eventually collapse.

⚠ **Scope.** All results are from ResNet-18 on Split CIFAR-100. Results establish existence proofs and quantitative baselines rather than universal laws.

## IX. Implications for Future Work

### A. The Manifold Anchor Regulariser

If the recovery manifold is stable at $k_t \approx 8$ and recoverability is governed by the orientation ($D_t$) of this manifold, the natural intervention is a regulariser that penalises rotation of the top-8 singular directions of the task-specific probe weight matrix. After learning task 0, extract $U_0 \in \mathbb{R}^{d \times 8}$—the top-8 right singular vectors of $W_0$. For each subsequent task t, add the anchor loss:

$L_{anchor} = \lambda \cdot \|U_0 - U_t\|^{2M}$

This penalises rotation of the recovery manifold while imposing no constraint on the remaining 504 dimensions, allowing full plasticity for new tasks.

**Paper 3 Central Question:** Can active preservation of the 8-dimensional recovery manifold—through targeted regularization of probe weight singular directions—maintain recoverability throughout continual learning without inhibiting plasticity for new tasks?

### B. Generalisation to Other Architectures

Testing whether $k_t \approx 8$ holds on ViT, MLP-Mixer, or plain convolutional networks would determine whether this quantity is an architecture-agnostic characteristic of the forgetting process or specific to residual networks.

### C. Practical Recovery Systems

Storing only $8 \times d = 4{,}096$ parameters per task could maintain near-full recoverability across the entire sequence—far more efficient than full checkpoint storage or replay buffers.

## X. Conclusion

This paper presented six experiments investigating the geometric structure and temporal dynamics of recoverability in continual learning. The central negative result—falsification of the Recoverability Diffusion hypothesis—was accompanied by the stronger positive finding: the Stable Recovery Manifold hypothesis, positing that forgotten task knowledge is preserved within a compact, stable, approximately 8-dimensional subspace, and that recoverability is governed by the *orientation* of this subspace rather than its information content.

The key quantitative results are: $k_t \approx 8.0$ ($\sigma = 0.82$) across all ten tasks; $R^2 = 0.82$ for a linear predictive model using three geometric variables; Pearson $r = -0.862$ for principal-angle drift as the dominant predictor; and a novel depth-stratification result confirming early-layer universalisation and late-layer specialisation as mechanistic complements.

Catastrophic forgetting, under these findings, is a geometric accessibility failure—a problem of representational orientation, not representational destruction. The dominant bottleneck is manifold orientation preservation. This reframing, if it holds under architectural and task-scale generalisation, represents a meaningful shift in how the continual learning community should think about what forgetting destroys and what intervention strategies should therefore target.